\documentclass[10pt,twocolumn,letterpaper]{article}

\usepackage{cvpr}              
\usepackage{multirow}
\usepackage{booktabs}
\usepackage{rotating}
\usepackage{graphicx}
\usepackage{colortbl}

\usepackage[dvipsnames]{xcolor}

\definecolor{cvprblue}{rgb}{0.21,0.49,0.74}
\usepackage[pagebackref,breaklinks,colorlinks,citecolor=cvprblue]{hyperref}

\def\paperID{11325} 
\def\confName{CVPR}
\def\confYear{2024}

\title{Multi-Object Tracking in the Dark}

\author{
Xinzhe Wang \quad Kang Ma \quad Qiankun Liu \quad Yunhao Zou \quad Ying Fu \footnotemark[1] \\
Beijing Institute of Technology \\
{\tt\small \{wangxinzhe, makang, liuqk3, zouyunhao, fuying\}@bit.edu.cn}}

\begin{document}
\maketitle

\renewcommand{\thefootnote}{\fnsymbol{footnote}}
\footnotetext[1]{Corresponding Author}

\renewcommand{\thefootnote}{\arabic{footnote}}


\begin{abstract}
Low-light scenes are prevalent in real-world applications (\eg autonomous driving and surveillance at night). Recently, multi-object tracking in various practical use cases have received much attention, but multi-object tracking in dark scenes is rarely considered. In this paper, we focus on multi-object tracking in dark scenes. To address the lack of datasets, we first build a \textbf{L}ow-light \textbf{M}ulti-\textbf{O}bject \textbf{T}racking  (\textbf{LMOT}) dataset. LMOT provides well-aligned low-light video pairs captured by our dual-camera system, and high-quality multi-object tracking annotations for all videos. Then, we propose a low-light multi-object tracking method, termed as \textbf{LTrack}.  We introduce the adaptive low-pass downsample module to enhance low-frequency components of images outside the sensor noises. The degradation suppression learning strategy enables the model to learn invariant information under noise disturbance and image quality degradation. These components improve the robustness of multi-object tracking in dark scenes. We conducted a comprehensive analysis of our LMOT dataset and proposed LTrack. Experimental results demonstrate the superiority of the proposed method and its competitiveness in real night low-light scenes. Dataset and Code: \url{https://github.com/ying-fu/LMOT}
\end{abstract}

\section{Introduction}
\label{sec:intro}

Multi-object tracking (MOT) aims to locate and associate multiple objects in video sequences. It is widely used in many downstream applications, such as video recognition \cite{choi2012unified,ma2023dynamic}, autonomous driving \cite{geiger2012we}, and surveillance \cite{ellis2010pets2010}. Recently, multi-object tracking in various practical use cases has garnered much attention \cite{milan2016mot16, dendorfer2020mot20,sun2022dancetrack,geiger2012we,cui2023sportsmot}, greatly advancing the development of MOT. However, these works are primarily tailored for high-quality inputs and overlook the prevalent low-light scenes in real-world scenes. Motivated by this, we study multi-object tracking in dark scenes.

Due to the physical limitations of existing cameras, acquiring high-quality videos under low-light conditions is difficult. One inherent difficulty in capturing consecutive video frames under such conditions is avoiding motion blur. Current camera technology typically requires short exposure times (usually just a few tens of milliseconds), but in low-light scenarios, the sensor struggles to capture an adequate number of photons within this limited duration. This limitation inevitably leads to degradation in image quality accompanied by higher noise levels. This presents two main challenges for MOT under low-light conditions. The first challenge is for collecting a low-light multi-object tracking dataset. Collecting and annotating a low-light MOT dataset is difficult and expensive. MOT requires dynamic object videos, but videos captured in low-light scenes have extremely low brightness, making it hard to recognize and annotate objects in the videos. The second challenge revolves around low-light multi-object tracking. The popular \textit{tracking-by-detection} paradigm \cite{liu2020gsm,liu2022online,cao2023observation,du2023strongsort,zhang2022bytetrack} generally consists of detector, motion-based association module, and appearance-based association module. These modules typically require high-quality input images. The poor quality of low-light images leads to severe performance degradation for both detectors and appearance-based correlation modules. A simple approach is to cascade low-light enhance modules \cite{cai2023retinexformer,jin2023dnf,wang2021seeing,jiang2019learning}, but this introduces additional computational costs. Furthermore, images optimized for visual quality may be suboptimal for downstream tasks \cite{chen2023instance,hong2021crafting,lee2023human}.

In this paper, we build a low-light multi-object tracking dataset (LMOT), specifically designed to address the challenges of multi-object tracking in dark scenes. To this end, we develop a dual-camera system that simultaneously captures well-lit and low-light video frames. The video pairs are highly aligned in both spatial and temporal dimensions, offering two key benefits. First, it enables us to annotate on the well-lit videos, resulting in high-quality annotations. Second, the well-lit videos can provide additional supervision information during the training phase, and strongly enhance the performance in the dark scenes. After careful annotation, we collect 32 video sequences (2.3$\times$ MOT17), over 35K frames (3.1$\times$ MOT17), and over 815K bounding boxes (2.8$\times$ MOT17). The RAW data is the output of the image sensor and is the input data of the image signal processor (ISP). It saves all information from the image sensor, which is crucial for capturing object information in dark scenes \cite{zou2023rawhdr,xu2023toward,chen2023instance}. Therefore, we collect RAW videos for LMOT.

Additionally, we propose a low-light multi-object tracking method, termed as LTrack. The low-light video is characterized by substantial sensor noise and poor image quality, which significantly degrades both shallow and deep feature representations, leading to reduced tracking performance. We observe that the sensor noise in low-light images exhibits a similarity to adversarial attacks \cite{szegedy2013intriguing,moosavi2017universal}. To address this issue, \textit{our main idea is to learn the invariant semantic information under noise disturbance quality degradation}. We present the adaptive low-pass downsample module (ALD). It employs spatial low-pass convolution to extract low-frequency components from images, excluding noises, and adaptively enhance the feature maps. We also present the degradation suppression learning strategy (DSL), which utilizes paired low-light videos to help the model suppress the noise disturbance and encourage image content response in the feature domain. We conduct a comprehensive analysis of our LMOT dataset and validate the superiority of our LTrack in real-world night scenes.

In summary, our main contributions are as follows:


\begin{itemize}
    \item We build the first low-light multi-object tracking dataset using a carefully constructed dual-camera system. It provides well-aligned low-light videos in RAW format, and high-quality MOT annotations for all videos.

    \item We propose a low-light multi-object tracking method. It utilizes the adaptive low-pass downsample module and the degradation suppression learning strategy to learn to extract invariant features from low-light videos.

    \item We conduct a comprehensive analysis of our dataset and the proposed method. Experimental results demonstrate the superiority of the proposed method and its competitiveness on real night scenes.
\end{itemize}

\section{Related Work}

In this section, we first review the current research status of low-light enhancement and low-light datasets. Then, we summarized the present research for multi-object tracking and tracking in the dark scenes.

\label{sec:related}
\vspace{0.5em}
\noindent \textbf{Low-light enhancement.} Traditional methods for low-light enhancement are primarily based on histogram equalization and Retinex theory \cite{guo2016lime,ibrahim2007brightness,jobson1997properties}. Recently, deep learning has been explored for many low-level tasks \cite{tian2022plug,tian2023local,wei2022tfpnp,zhang2021hyperspectral,fu2021residual,zou2022estimating,zhang2022deep,zou2021learning}, and achieved superior results on low-light enhancement \cite{cai2023retinexformer,chen2018learning,guo2020zero,fu2022gan,sharma2021nighttime,jin2023dnf,fu2023dancing,xu2022snr}. While these methods are capable of recovering images with high visual quality, they often require heavy computation and may not consider downstream tasks, resulting in suboptimal performance. In contrast, our approach focuses on directly learning multi-object tracking from low-light images, thus bypassing the low-light enhancement.

\vspace{0.5em}
\noindent \textbf{Low-light datasets.} 
\label{sec:related_low_light_datasets}
The long-short exposure is a widely used method to collect paired low-light images, but can only be used to collect low-light images for static scenes. \cite{chen2018learning,wei2018deep,chen2019seeing}. To capture dynamic scenes, some works designed the mechatronic system. They obtain paired low-light data by repeating the motion twice \cite{wang2021seeing,fu2023dancing}. However, these mechatronic systems cannot be used to collect dynamic object videos in the wild. Jiang \textit{et al.} \cite{jiang2019learning} designed a dual-camera system that simultaneously captures paired well-lit and low-light videos, making it possible to capture dynamic scenes and dynamic object videos for multi-object tracking. Zou \textit{et al.} \cite{zhang2021learning} setup an optical system to collect paired videos and event streams. These works explore various ways to construct low-light datasets and inspire research on high-level vision tasks in dark scenes, such as LOD \cite{hong2021crafting} for object detection, LIS \cite{chen2023instance} for instance segmentation, and ExPose \cite{lee2023human} for human pose estimation. These datasets provide paired low-light data only for image tasks, and cannot be expanded for multi-object tracking in dark scenes.

\begin{figure}
    \setlength{\abovecaptionskip}{2pt}
    \centering
    \includegraphics[width=1\linewidth]{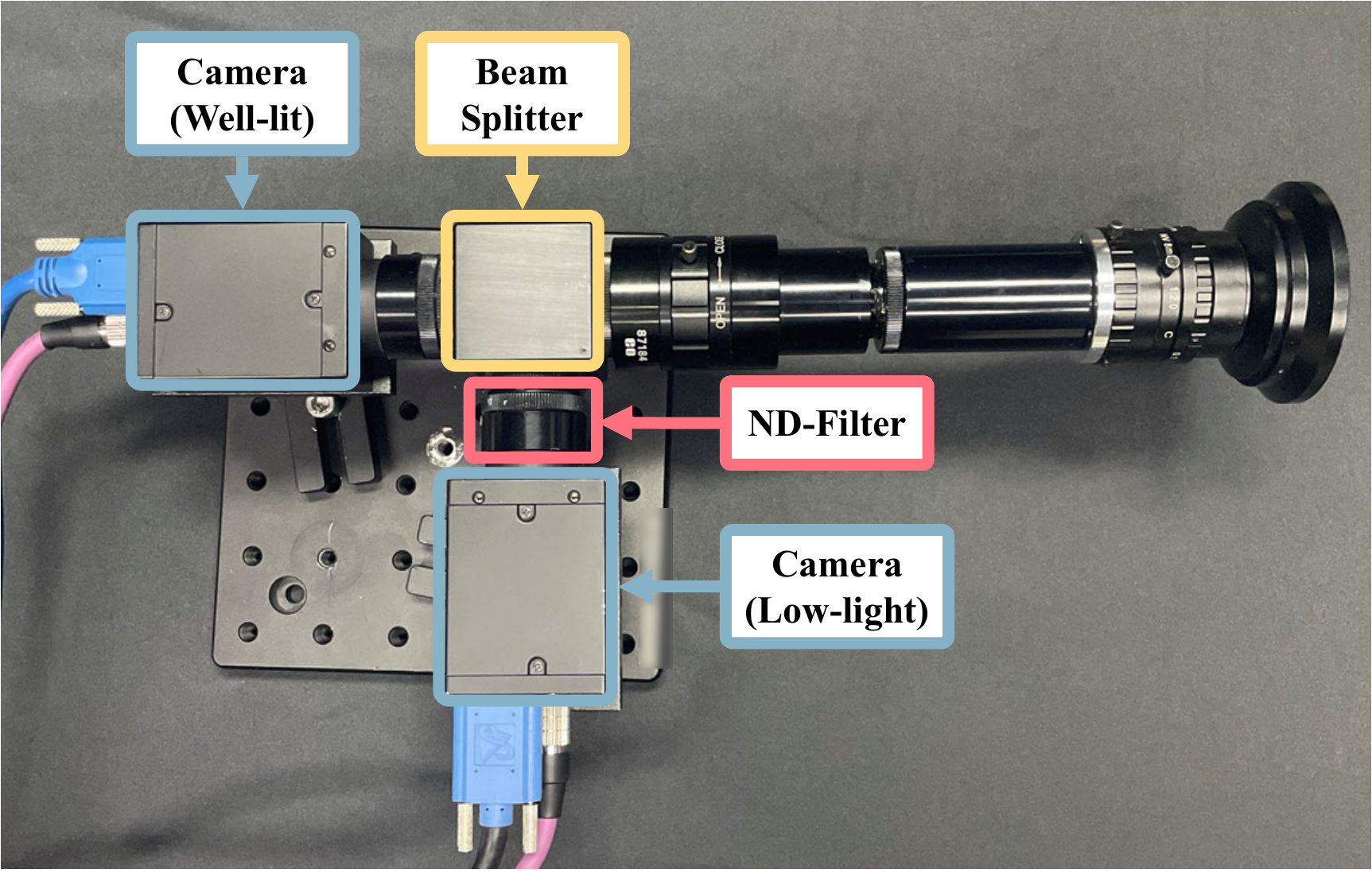}
    \caption{Our dual-camera system. It consists of two cameras, a beam splitter, and an ND-filter. Two cameras of identical models are meticulously engineered to achieve pixel-by-pixel alignment in the captured video data.}
    \label{fig:dual_cam_img}
    \vspace{-1em}
\end{figure}

\begin{figure*}
    \setlength{\abovecaptionskip}{2pt}
    \centering
    \raisebox{4mm}{\rotatebox{90}{\fontsize{6}{0}\selectfont Well-lit \hspace{4mm} Scaled Low-light \hspace{3mm} Low-light}}
    \includegraphics[width=0.98\linewidth]{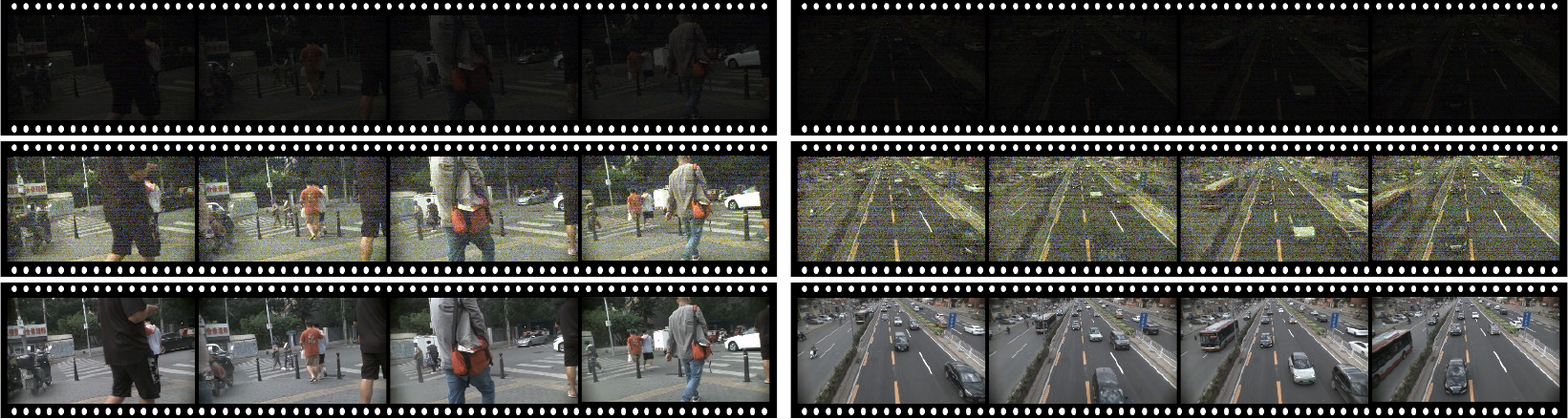} 
    \caption{Two example videos from our LMOT dataset. It provides well-aligned low-light video pairs and MOT annotations for all videos. The time interval between adjacent frames is $1s$. The first row is the low-light video, the second row is the scaled low-light video and the last row is the well-lit video. Our LMOT dataset is collected from city outdoor scenes.}
    \label{fig:data_preview}
    \vspace{-1em}
\end{figure*}

\vspace{0.5em}
\noindent \textbf{Multi-object tracking datasets.} MOT15 \cite{leal2015motchallenge} is the first large-scale benchmark for multi-object tracking. MOT17 \cite{milan2016mot16}  stands as one of the most widely applied MOT benchmarks. MOT20 \cite{dendorfer2020mot20} focuses on very crowded scenes. These three datasets are for pedestrians. KITTI \cite{geiger2012we} and BDD100K \cite{yu2020bdd100k} are for autonomous driving scenarios.  DanceTrack \cite{sun2022dancetrack} focuses on dancing scenes and is characterized by similar appearance and diverse motions. Recently, SportsMOT \cite{cui2023sportsmot} aims to track athletes and encourage algorithms to promote both appearance and motion association. These datasets explore multi-object tracking in various practical use cases, but none of them consider multi-object tracking in dark scenes.

\vspace{0.5em}
\noindent \textbf{Object tracking in the dark.} To track in low-light scenes, some methods explore to use multi-modal information for single-object tracking, such as event camera \cite{zhang2021object,zhu2023cross}, depth \cite{song2013tracking,yang2023resource} and thermal \cite{wang2020cross,zhang2022visible} devices. Park \textit{et al.} \cite{park2023multi} proposed to use Short-Wave Infrared (SWIR) images for multi-object tracking, since its advantages in terms of robustness in low-light conditions. The common drawback of these methods is that they require additional hardware equipment and cannot be applied to the most widely used CMOS imaging systems. 

SORT \cite{bewley2016simple} uses the Kalman Filter motion model and employs IoU for association. ByteTrack \cite{zhang2022bytetrack} enhances tracking performance by considering low-confidence bounding boxes. OS-SORT \cite{cao2023observation} enhances SORT by restoring lost targets. Recently, Transformer has been explored for MOT \cite{gao2023memotr,sun2020transtrack,meinhardt2022trackformer,zeng2022motr,zhang2023motrv2}. These methods have achieved high performance in many practical scenarios, but they do not consider dealing with low-light conditions. We focus on multi-object tracking under low-light conditions. Based on RAW videos, our method is highly practical with excellent performance and does not require additional hardware.

\begin{table}
    \setlength{\abovecaptionskip}{2pt}
    \setlength{\tabcolsep}{3pt}
    \centering
    \resizebox{1\linewidth}{!}{
        \begin{tabular}{lcccccc}
            \toprule
            Dataset & Format & Videos & Frames & Length (s)  & Bbox & Tracks \\
            \midrule
            MOT17 \cite{milan2016mot16} & sRGB &  14 & 11,235 & 463 & 292,733 & 1,342 \\
            MOT20 \cite{dendorfer2020mot20} & sRGB &  8 & 13,410 & 535 & 1,652,040 & 3,456 \\
            DanceTrack \cite{sun2022dancetrack} & sRGB &  100 & 105,855 & 5,292 & - & 990 \\
            SportMOT \cite{cui2023sportsmot} & sRGB & 240 & 150,379 & 6,015 & 1,629,490 & 3,401 \\
            \midrule
            KITTI \cite{geiger2012we} & sRGB & 21 & 8,000 & - & 47,000 & 917 \\
            BDD100K \cite{yu2020bdd100k} & sRGB & 1,600 & 318,000 & - & 3,300,000 & 131,000 \\ 
            SWIR \cite{park2023multi} & SWIR &  - & 7,309 & - & 57,221 & - \\
            \midrule
            LMOT & RAW &  32 & 35,120 & 1,756 & 815,550 & 4,090 \\
            \bottomrule
        \end{tabular}
    }
    \caption{Comparison of statistics between existing MOT datasets and our LMOT dataset}
    \label{tab:data_statistic_compare}
    \vspace{-1em}
\end{table}

\section{Low-light Multi-object Tracking Dataset}
\label{sec:dataset}

In this section, we first introduce our dual camera system and the details of collecting and annotating our low-light multi-object tracking (LMOT) dataset. Then, we analyzed the statistical characteristics of our LMOT dataset.

\begin{figure*}
\setlength{\abovecaptionskip}{2pt}
\centering
\resizebox{1\linewidth}{!}{
    \begin{subfigure}{0.355\linewidth}
        \includegraphics[width=1\linewidth]{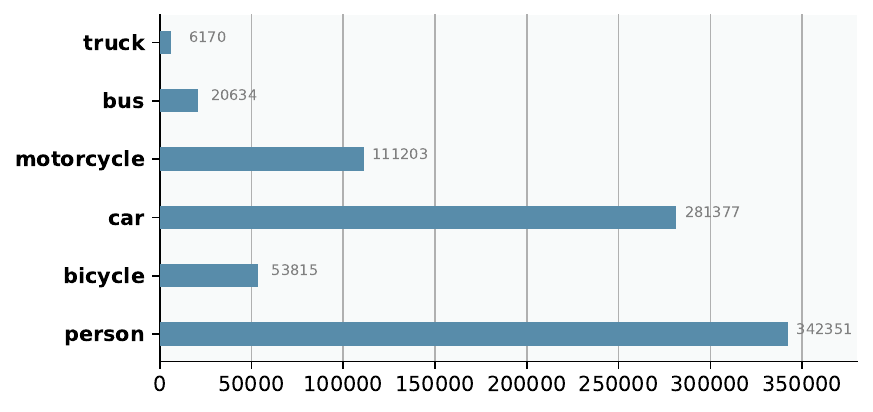}
        \caption{Number of instances per category}
        \label{fig:data_statics_category}
    \end{subfigure}
    \begin{subfigure}{0.28\linewidth}
        \includegraphics[width=1\linewidth]{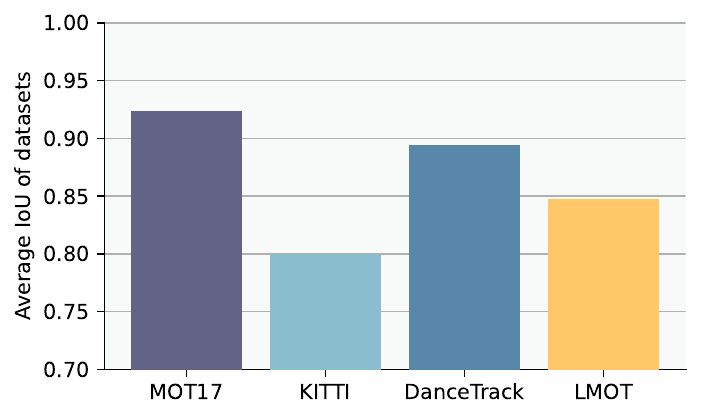}
        \caption{IoU on adjacent frames}
        \label{fig:data_statics_iou}
    \end{subfigure}
    \begin{subfigure}{0.332\linewidth}
        \includegraphics[width=1\linewidth]{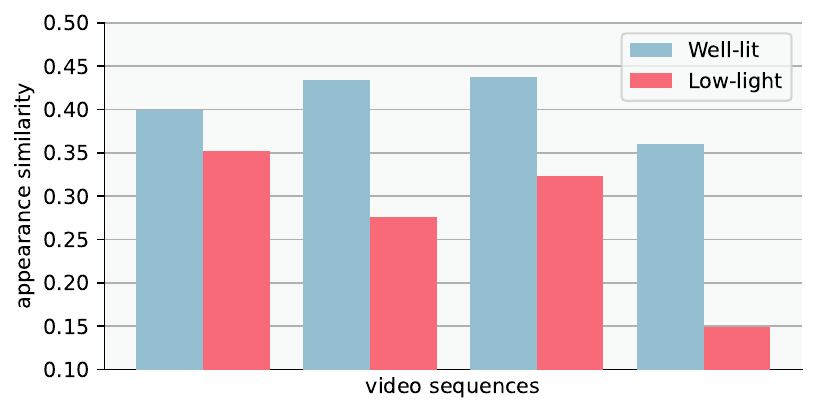}
        \caption{Cosine distance of appearance features}
        \label{fig:data_statics_cos}
    \end{subfigure}
}
\caption{(a) Number of instances per category. LMOT consists of 6 categories, most of the instances are the person and car. (b) IoU on adjacent frames. Compared to MOT17, KITTI, and DanceTrack, LMOT has a roughly average score. This indicates that LMOT has a relatively normal movement speed. (c) Cosine distance of appearance features. The cosine distance is smaller under low-light conditions, indicating that the appearance distinguishability is decreased under low-light conditions.}
\label{fig:data_statics}
\vspace{-1em}
\end{figure*}

\subsection{Dataset Construction}

Multi-object tracking requires dynamic scenes and object video. To collect low-light videos for multi-object tracking, we build a dual-camera system \cite{jiang2019learning}, which is illustrated in \cref{fig:dual_cam_img}. It can simultaneously capture paired low-light and well-lit video pairs. Its main components include a beam splitter, a neutral density (ND) filter, and two \textit{FLIR Grasshopper3 GS3-U3-23S6C} cameras. The beam splitter divides the incoming light into two separate paths. This arrangement allows one camera to capture well-lit images directly, while the other camera records low-light images, with the ND-filter attenuating the light intensity. To ensure temporal synchronization of the video frames, we employ the hardware interface to trigger the camera exposure events. Moreover, to avoid frame loss, our dual-camera system uses two independent hardware interfaces for data transmission and is equipped with a high-speed solid-state drive. Thanks to precise calibration, our dual-camera system can capture paired low-light and well-lit videos in real time for dynamic scenes and objects. More details about our dual-camera system are given in \textit{supplementary materials}.

\begin{table}[t]
\setlength{\abovecaptionskip}{2pt}
\centering
\resizebox{1\linewidth}{!}{
    \begin{tabular}{lccccc}
    \toprule
     Dataset & Split & Videos & Bbox & Tracks & Paired Well-lit \\
    \midrule
    \multirow{3}*{LMOT-dual} & train & 11 & 309,466 & 1533 & \checkmark \\
     & val & 4 & 131,781 & 626 & \checkmark \\
     & test & 11 & 312,742 & 1644 & \checkmark\\
     \midrule
    LMOT-real & real & 6 & 61,561 & 287 \\
    \bottomrule
    \end{tabular}
}
\caption{Detailed statistics and data splits for LMOT.}
\label{tab:data_statistic_split}
\vspace{-1em}
\end{table}

We save the video frames in RAW format before the images are processed by the camera image signal processor (ISP). In terms of camera settings, we set the exposure time for both cameras to $10ms$, and the frame rate is fixed at 20. This setting is feasible to avoid motion blur. We adjust the gain level for well-lit cameras to achieve optimal image quality. The gain for low-light cameras is consistently set to the maximum value, to simulate low-light capturing setup in real scenarios. We also collect a real low-light MOT dataset (LMOT-real) to evaluate performance in real-world dark scenes. These videos are captured using a single camera with the same camera settings.

Our LMOT dataset contains a variety of city outdoor scenes, including roads, overpasses, pedestrians, and intersection. The overpass scenes take an overhead shot of objects, while all other scenes are captured from the perspective of pedestrians. To account for the impact of camera motion, we introduce arbitrary horizontal rotations and vertical random movements to the camera. \cref{fig:data_preview} shows two sampled video sequences from our LMOT dataset.

We annotate six types of moving objects, including car, person, bicycle, motorcycle, bus, and truck. The annotated labels include bounding boxes, identifications, and visibility status. For partly occluded objects, a full box is annotated. For a fully occluded object, an estimated box is annotated. Each object has a unique ID throughout the entire video. Thanks to well-aligned low-light and well-lit videos, we can annotate well-lit videos to simultaneously obtain labels for low-light videos. This greatly reduces the annotation difficulty and enhances quality. Lastly, we carefully review all the annotation results.

\subsection{Dataset Statistic} 
LMOT is a large-scale dataset that focuses on multi-object tracking in dark scenes. We compare the statistics of LMOT with existing MOT datasets in \cref{tab:data_statistic_compare}. It can be seen that LMOT is approximately three times larger than MOT17 \cite{milan2016mot16}. Compared to large-scale datasets such as DanceTrack \cite{sun2022dancetrack}, SportsMOT \cite{jobson1997properties}, KITTI \cite{geiger2012we} and BDD100K \cite{yu2020bdd100k}, the scale of LMOT is still considerable. It should be emphasized that these datasets are not for multi-object tracking in dark scenes and only provide sRGB images. Compared to SWIR, our LMOT dataset has approximately $5\times$ frames and $14\times$ bounding boxes. The detailed statistics and data splits for LMOT are shown in \cref{tab:data_statistic_split}.

We present the number of instances for each category in \cref{fig:data_statics} ({\color{red}a}). The majority of instances are persons and cars. As shown in \cref{fig:data_statics} ({\color{red}b}), the average IoU on adjacent frames of LMOT is lower than MOT17 and DanceTrack, but higher than KITTI. This indicates that the motions in LMOT are fast but within normal range. Following \cite{sun2022dancetrack}, we use cosine distance of appearance features \footnote{We use UniTrack \cite{wang2021different} to extract appearance features.} to evaluate the appearance similarity. From \cref{fig:data_statics} ({\color{red}c}), we can see that the cosine distance of appearance features under low-light conditions is smaller than that under well-lit conditions. In other words, the appearance of objects will deteriorate under low-light conditions, making them harder to distinguish.

\section{Low-light Multi-object Tracking}
\label{sec:method}
\begin{figure*}
\setlength{\abovecaptionskip}{2pt}
\centering
\includegraphics[width=1\linewidth]{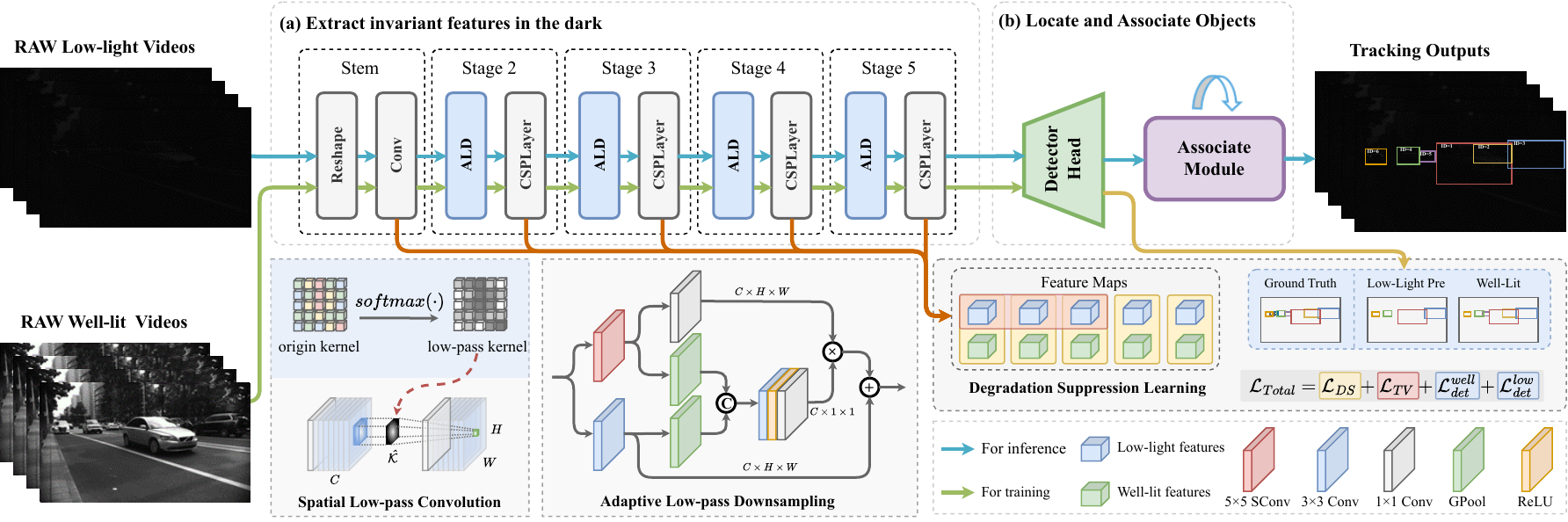}
\caption{The overall framework of the proposed low-light multi-object tracking method, termed as \textbf{LTrack}. It employs adaptive low-pass downsample module and degradation suppression learning strategy, enabling the model to learn invariant features from low-light videos.}
\label{fig:method_main_framework}
\vspace{-1em}
\end{figure*}

In this section, we peresnt our low-light multi-object tracking method (LTrack). Our main idea is to \textit{learn the invariant semantic information under noise disturbance quality degradation}. The overall framework can be seen in \cref{fig:method_main_framework}.

\subsection{Formulation and Motivation}

In low-light scenes, the camera can capture only a small number of photons in a single exposure. Thus, the potential sensor noise is highlighted, resulting in significant degradation to the image quality \cite{wei2020physics}. We observed that directly feeding low-light images to the network without any special design leads to the feature map degradation and significantly reduces the performance of the model (see in \cref{sec:sub_analysis_low_light}). A direct solution is to apply low-light enhancement techniques, which focus on learning a mapping function from low-light images to clean well-lit images. Since it is a highly ill-posed problem, learning such a mapping function requires considerable computing and storage overhead. Although DNN-based methods have achieved excellent performance in low-light enhancement, images enhanced for visual quality may be suboptimal for downstream tasks.

In this work, we perform multi-object tracking from low-light images, bypassing the low-light enhancement. Leveraging RAW videos, the network obtains more original scene information compared to sRGB. To enhance the performance and robustness of the multi-object tracking model, our main idea is to earn the invariant semantic information under noise disturbance quality degradation. Thus, we present the adaptive low-pass downsampling (ALD) module to enhance the low-frequency components of images and filter out high-frequency noise. We also propose the degradation suppression learning strategy (DSL), which utilizes paired low-light videos to help the model suppress image noise disturbance and encourage image content response in the feature domain.

\subsection{Adaptive Low-pass Downsampling}
The downsampling operation reduces the feature size while preserving the most important information. The noise in low-light images introduces high-frequency disturbance to the feature maps, which can mislead the preservation of object information. To weaken the impact of high-frequency noises and enhance the low-frequency part of the feature map, we introduce spatial low-pass convolution (SConv) to extract low-frequency features from the noised feature maps. The softmax function is used to constrain the original convolution kernel to be low-pass as
{\setlength{\abovedisplayskip}{3pt}
\setlength{\belowdisplayskip}{3pt}
\begin{align}
    \mathcal{\hat{K}}_{i,j}=\frac{exp(\mathcal{K}_{i,j})}{\sum_{p,q}{exp(\mathcal{K}_{p,q})}}
\end{align}}
where $\mathcal{K}$ and $\mathbf{\mathcal{\hat{K}}}$ are the origin convolution kernel and low-pass convolution kernel. We initialize the spatial low-pass convolution kernel using a standard Gaussian kernel. The kernel size is set to $5$ to obtain more spatial information. Then, the obtained low-frequency features are adaptively weighted and fused into the original features. We use global average pooling to obtain channel descriptors for both origin features and low-frequency features. These descriptors are then fed into a fully connected layer to compute the weight values.

\subsection{Degradation Suppression Learning}
\begin{figure}
\setlength{\abovecaptionskip}{2pt}
\centering
\raisebox{7.5mm}{\rotatebox{90}{\fontsize{6}{0}\selectfont Low-light \hspace{7mm} Well-lit}}
\begin{subfigure}{0.315\linewidth}
    \includegraphics[width=1\linewidth]{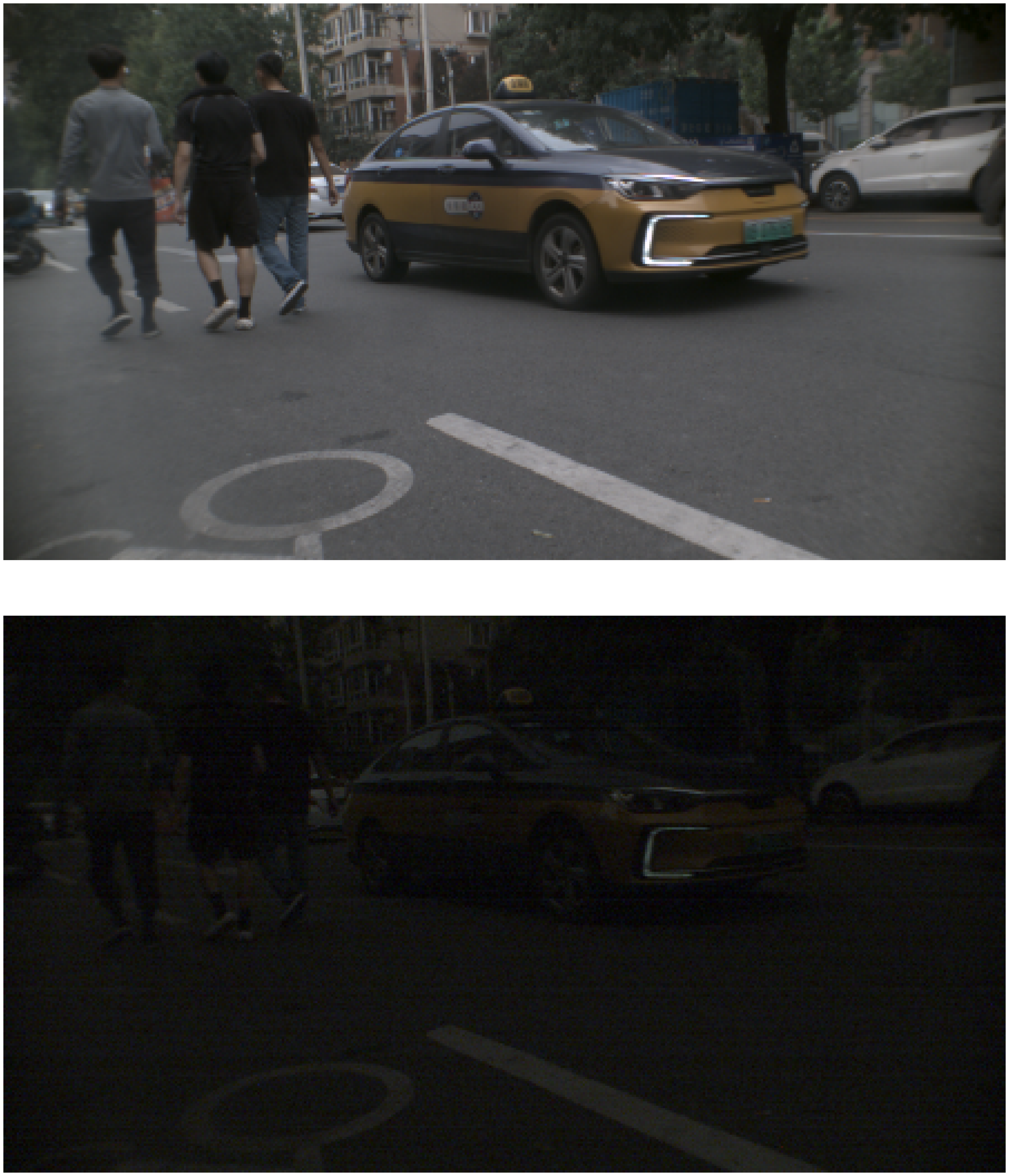}
    \caption{Input}
\end{subfigure}
\begin{subfigure}{0.315\linewidth}
    \includegraphics[width=1\linewidth]{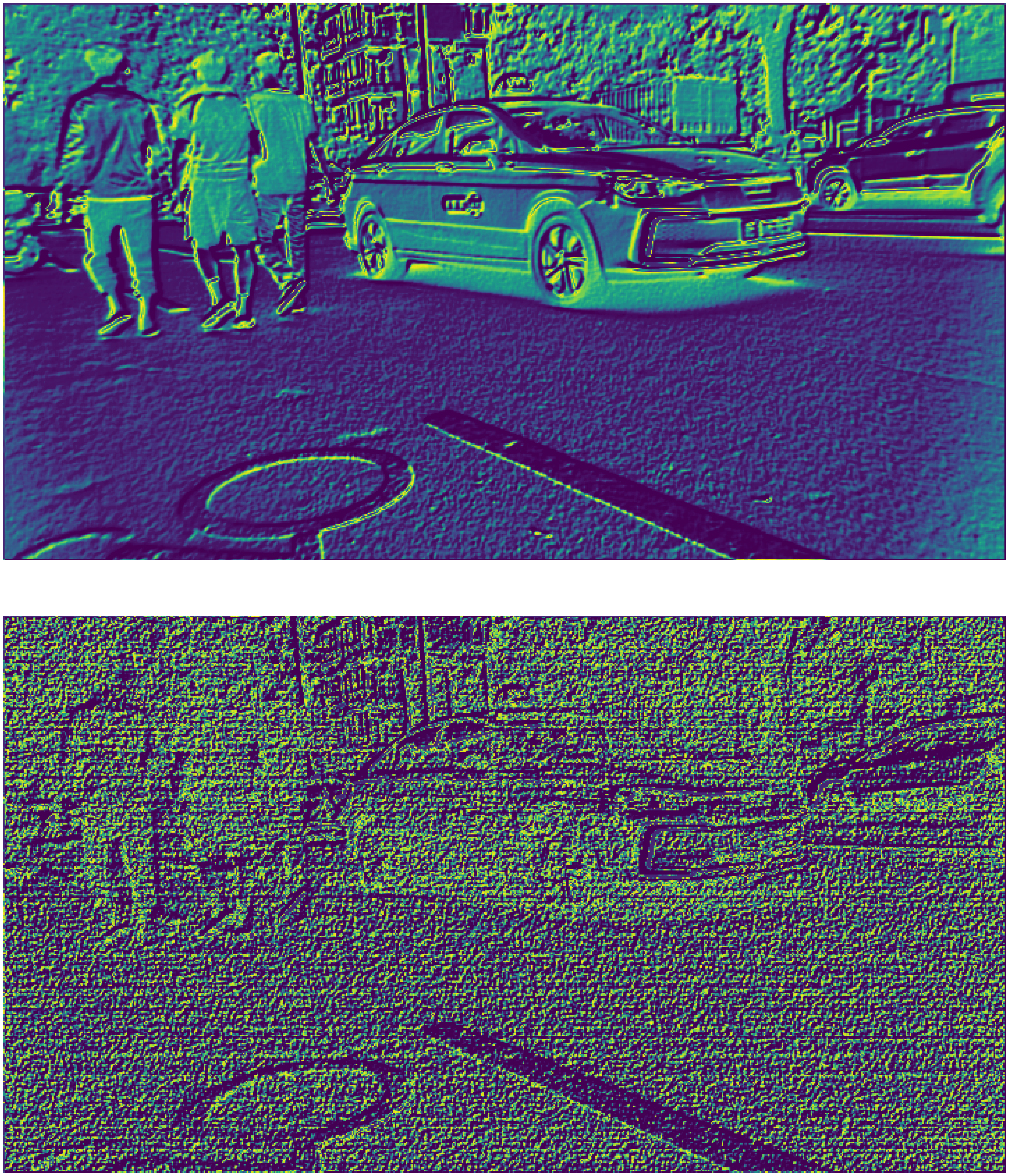}
    \caption{Shallow feature}
\end{subfigure}
\begin{subfigure}{0.315\linewidth}
    \includegraphics[width=1\linewidth]{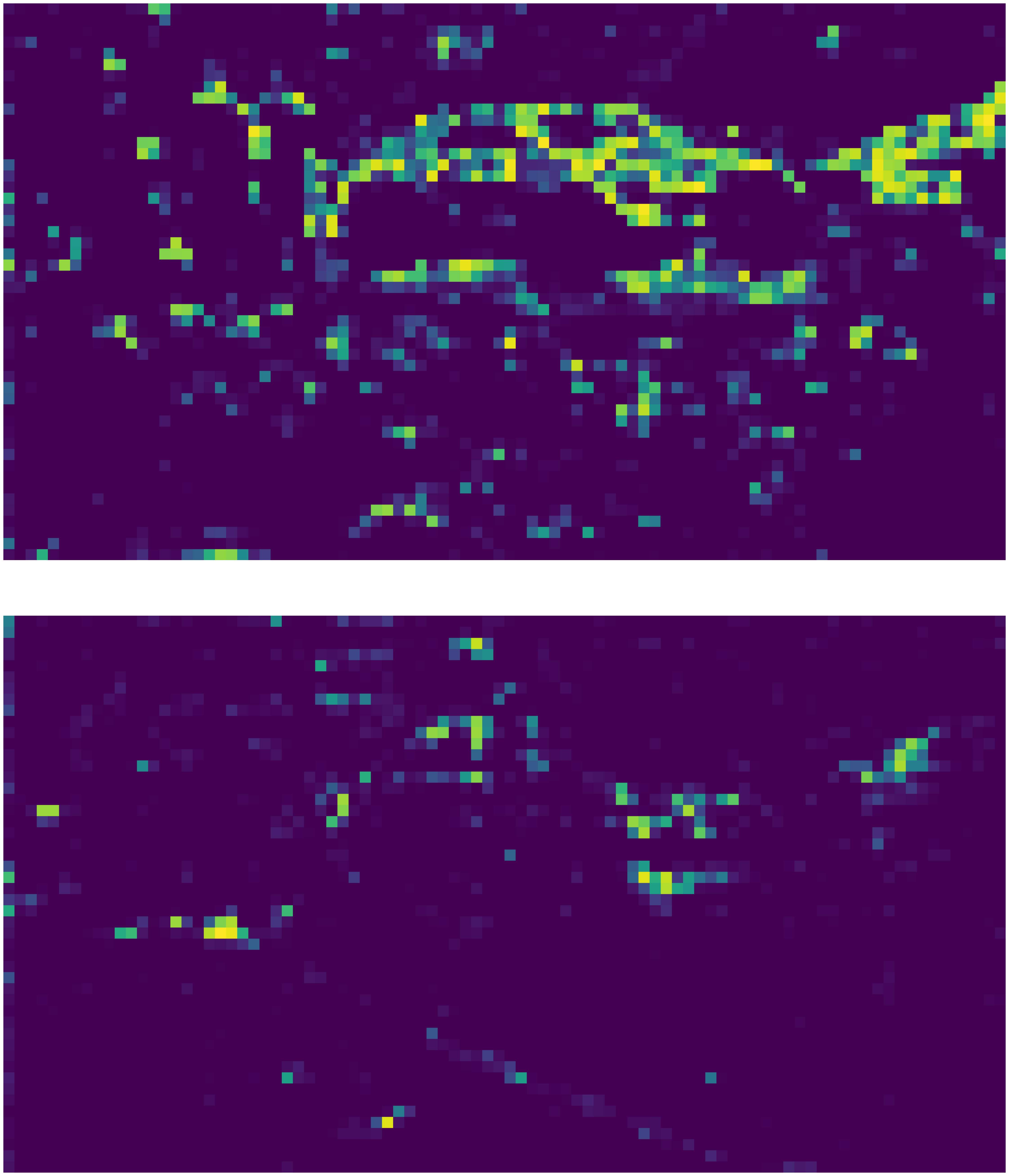}
    \caption{Deep feature}
\end{subfigure}
\caption{Visualization of shallow and deep features for well-lit and low-light images. It can be seen that, under low-light conditions, the shallow feature is full of noise, and the deep feature exhibits lower responses to objects.}
\label{fig:exp_analysis_feature_visual}
\vspace{-1em}
\end{figure}

Given that a low-light image and well-lit image pair share the same content, the model should exhibit identical feature responses to them. However, the low-light image results in shallow features full of noise, and the deep feature exhibits lower responses to objects (as shown in \cref{fig:exp_analysis_feature_visual}). To address this, our idea is to suppress image noise in shallow features and use well-lit images to help model learning disturbance invariant information from low-light images. The degradation suppression loss can be expressed as
{\setlength{\abovedisplayskip}{3pt}
\setlength{\belowdisplayskip}{3pt}
\begin{align}
    \mathcal{L}_{DS} = \sum_{l}||\mathbf{F}^{well}_l-\mathbf{F}^{low}_l||^2_2
\end{align}}
where $\mathbf{F}^{well}_{l}$ and $\mathbf{F}^{low}_{l}$ denotes $l$-th feature map corresponding to well-lit and low-light image, respectively. To further suppress the noise in shallow features, we also introduce the Total Variation (TV) loss \cite{rudin1992nonlinear} to features as
{\setlength{\abovedisplayskip}{3pt}
\setlength{\belowdisplayskip}{3pt}
\begin{align}
    \mathcal{L}_{TV} = \sum_{l}||\mathbf{G}^{row}\mathbf{F}_{l}^{low}||^2_2+||\mathbf{G}^{col}\mathbf{F}_{l}^{low}||^2_2
\end{align}}
where $\mathbf{G}^{row}$ and $\mathbf{G}^{col}$ are the first derivative matrix to role and column. The TV loss adds spatial smoothing constraints to features, which helps model learning to extract noise invariant features. Both the well-lit and low-light images are used to train the model by a common detection loss ${L}_{det}(\cdot)$. The total loss is $\mathcal{L}_{Total}$, \ie,
{\setlength{\abovedisplayskip}{3pt}
\setlength{\belowdisplayskip}{3pt}
\begin{align}
    \mathcal{L}_{Total} = \mathcal{L}_{det}^{well} + \alpha\mathcal{L}_{det}^{low} +  \beta \mathcal{L}_{DS} + \gamma\mathcal{L}_{TV}
\end{align}}
where is the $\mathcal{L}_{det}^{well}$ and $\mathcal{L}_{det}^{low}$ are the detection loss for well-lit and low-light images, $\alpha$ and $\beta$ are loss weights. We set $\alpha$, $\beta$ and $\gamma$ to $1$, $1$ and $0.01$, respectively.

\section{Experiments}
\label{sec:experiments}

\subsection{Experiment Setup}

\noindent \textbf{Dataset Split.} In structuring our dataset, we randomly split the videos into training, validation, and testing sets, consisting of 11, 4, and 11 videos respectively. We also provide LMOT-real dataset that is captured in real-world night scenes with 6 videos. Detailed dataset split and statistical information are shown in \cref{tab:data_statistic_split}.

\noindent \textbf{Evaluation Metrics.} Following \cite{sun2022dancetrack,cui2023sportsmot}, we recommend using HOTA \cite{luiten2021hota} as the main evaluation metric to simultaneously evaluate the performance of detection and association. We also employ AssA and IDF1 \cite{ristani2016performance} to evaluate association performance, MOTA, and Deta to evaluate detection performance. There are two ways to combine metrics for all classes into a single score. One is by averaging metrics over the class values, and the other is by over the detection values. To avoid possible result bias caused by some categories with fewer samples (such as Bus and Truck), we combine scores by averaging over the detection values.

\vspace{0.5em}
\noindent \textbf{Implementation Details.} Following ByteTrack \cite{zhang2022bytetrack} and OC-SORT \cite{cao2023observation}, we use YOLOX \cite{ge2021yolox} as our detector. The trackers are pre-trained on the COCO \cite{lin2014microsoft} dataset and then trained on our LMOT dataset for 24 epochs. We apply data augmentation strategies including random flip, scale jitter of resizing, and Mosaic. We also use the physical-based noise model \cite{wei2020physics} for RAW image augment. We use SGD optimizer with weight decay $10^{-4}$ and cosine learning rate schedule, the initial learning rate is $10^{-4}$ and gradually reduces to $10^{-5}$. We apply linear interpolation as post-processing to all trackers, with the maximum gap set to 20.

\subsection{Analysis under low-light conditions}
\label{sec:sub_analysis_low_light}


We analyze the impact of low-light conditions on multi-object tracking using the LMOT validation set. It should be emphasized that the low-light and well-lit video pairs are perfectly aligned in LMOT.


\vspace{0.5em}
\noindent \textbf{Impact to detectors}. We first analyze the impact of lighting conditions on the detector. We select YOLOX as the detector since it is widely used in MOT areas \cite{zhang2022bytetrack,cao2023observation,cui2023sportsmot}. We train the detector using well-lit images (WL), low-light images (LL), and all the images (AL). Then, test them on both well-lit and low-light images. The results are shown in \cref{tab:exp_analysis_det}. From the table, we can see that the model trained by well-light images achieves the best result on well-light images, but its performance significantly decreased on low-light images. Further, we visualized the feature maps under these two lighting conditions in \cref{fig:exp_analysis_feature_visual}. It can be seen that both the shallow feature and deep features of the low-light image are significantly degraded due to the sensor noises. We also observed that using low-light images can largely improve the performance of the decoder under low-light conditions, and use all images to achieve the best performance. But this result is much lower than that under well-lit conditions.

\begin{table}
\centering
\setlength{\abovecaptionskip}{2pt}
\resizebox{1\linewidth}{!}{
    \begin{tabular}{cc|cccccc}
        \toprule
         \multicolumn{2}{c|}{Training data} & \multicolumn{3}{c}{Well-lit} & \multicolumn{3}{c}{Low-light} \\
         \midrule
         WL & LL & mAP & mAR & AP50 & mAP & mAR & AP50 \\
         \midrule
         \checkmark & & \textbf{37.0} & \textbf{45.0} & \textbf{65.5} & 3.1 &  4.8 & 6.6 \\
            & \checkmark & 23.0 & 30.8 & 40.8 & 16.9 & 23.7 & 30.3\\
         \checkmark & \checkmark & 28.7 & 35.2 & 49.1 & \textbf{17.9} & 2\textbf{4.1} & \textbf{32.2} \\
        \bottomrule
    \end{tabular}
}
\caption{Analysis on LMOT validation set for detector (YOLOX \cite{ge2021yolox}). WL and LL indicate the well-lit and low-light, respectively. It shows that it is hard to detect objects under low-light conditions.}
\label{tab:exp_analysis_det}
\vspace{-0.5em}
\end{table}

\begin{table}
\setlength{\abovecaptionskip}{2pt}
\centering
\resizebox{1\linewidth}{!}{
    \begin{tabular}{ccc|ccccc} 
        \toprule
         Cond. & Mot.  & App. & HOTA & AssA & IDF1 & MOTA & DetA   \\
         \midrule
         - & \checkmark &            & 86.1 & \textbf{79.5} & 72.9& 82.6& 93.4   \\
        WL & & \checkmark            & \textbf{87.3} & 77.1 & 82.5& \textbf{98.8}& \textbf{98.4}   \\
        WL & \checkmark & \checkmark & 83.7 & 78.0 & \textbf{85.6}& 96.8& 90.0   \\
        LL & & \checkmark            & 53.8 & 32.1 & 40.0& 67.3& 90.2   \\
        LL & \checkmark & \checkmark & 83.4 & 77.5 & 85.0& 96.0& 89.8   \\
        \bottomrule
    \end{tabular}
}
\caption{Analysis on LMOT validation set for different association models. Cond, Mot, and App indicate the light condition, motion, and appearance, respectively. The detection boxes are ground-truth boxes. It indicates that appearance information is effective for association under well-lit conditions, but dose not as effective under low-light conditions.}
\label{tab:exp_analysis_mot}
\vspace{-1em}
\end{table}

\begin{table*}
\setlength{\abovecaptionskip}{2pt}
\setlength{\tabcolsep}{5.5pt}
\centering
    \begin{tabular}{lcccccccccccccc}
    \toprule
    \multirow{2}*{Method} & \multicolumn{3}{c}{ByteTrack\cite{zhang2022bytetrack}} && \multicolumn{3}{c}{OC-SORT\cite{cao2023observation}} &&  \multicolumn{3}{c}{Detection} && \multirow{2}*{Params} &  \multirow{2}*{FLOPS}\\
    \cmidrule{2-4}
    \cmidrule{6-8}
    \cmidrule{10-12}
    &    HOTA & AssA & IDF1 && HOTA & AssA & IDF1 && mAP & AP50 & AP75 && (M) & (G) \\
    \midrule
     Base-low                           & 28.0 & \underline{42.9} & 10.9 && 28.8 & 40.3 & 34.3 && 18.8 & 32.4 & 18.9 && 99.00 & 793.29 \\
     Base-well                          & 13.5 & 36.8 & 32.9 && 13.8 & 31.7 & 12.0 && 5.5 & 9.7 & 5.5 && 99.00 & 793.29 \\
     Base-all                           & 28.1 & 42.6 & 32.9 && 28.9 & 39.8 & 34.6 && 19.0  & 32.5 & 19.1 && 99.00 & 793.29 \\
    \midrule
     LLFlow \cite{wang2022low}           & 28.8 & 42.9 & 34.0 && \underline{29.4} & \underline{41.0} & 35.0 && \underline{22.4} & \underline{38.3} & \underline{22.6} && 116.42 & 8755.91  \\
     SDSD \cite{wang2021seeing}          & \underline{29.1} & 41.1 & \textbf{35.5} && 29.3 & 36.9 & \textbf{36.8} && 19.8 & 36.2 & 19.3 && 103.30 & 1136.39 \\
     \midrule
     DNF \cite{jin2023dnf}               & 27.6 & 40.9 & 33.0 && 28.1 & 36.4 & 35.3 && 19.1 & 35.6 & 18.2 && 101.83 & 907.33 \\
     SMOID \cite{jiang2019learning}      & 28.0 & 39.7 & 33.9 && 28.7 & 36.7 & 35.6 && 19.1 & 35.5 & 18.2 && 120.88 & 7286.17  \\
    \midrule
     RAOD \cite{xu2023toward}            & 26.1 & 42.9 & 29.5 && 26.1 & \textbf{42.9} & 29.5 && 18.3 & 31.3 & 18.4 && 99.07 & 897.04  \\
     \midrule
     \rowcolor{gray!20} \textbf{LTrack (Ours)} & \textbf{29.4} & \textbf{43.2} & \underline{35.2} && \textbf{29.8} & 39.0 & \underline{36.7} && \textbf{23.4} & \textbf{40.6} & \textbf{23.5} && 100.43 & 800.92 \\
    \bottomrule
    \end{tabular}
\caption{Experimental results on LMOT dataset. Base-low, Base-well and Base-all indicate the baseline model trained on low-light, well-lit and all of them, respectively. The number of parameters and prediction latency of each method are reported along with the accuracy}
\label{tab:exp_main_dual}
\vspace{-1em}
\end{table*}

\begin{figure}
    \setlength{\abovecaptionskip}{2pt}
    \centering
    \begin{minipage}[]{0.49\linewidth}
        \begin{subfigure}{1\linewidth}
            \includegraphics[width=1\linewidth]{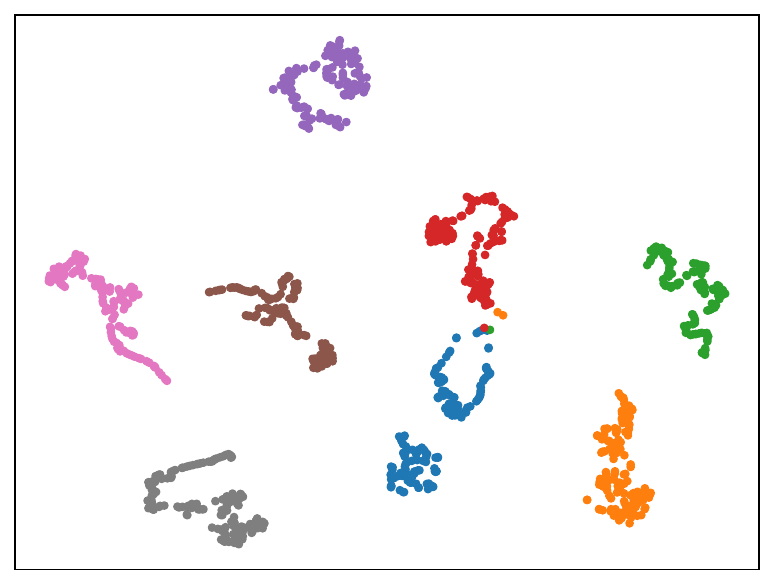}
            \caption{Well-lit}
        \end{subfigure}
    \end{minipage}
    \begin{minipage}[]{0.49\linewidth}
        \begin{subfigure}{1\linewidth}
            \includegraphics[width=1\linewidth]{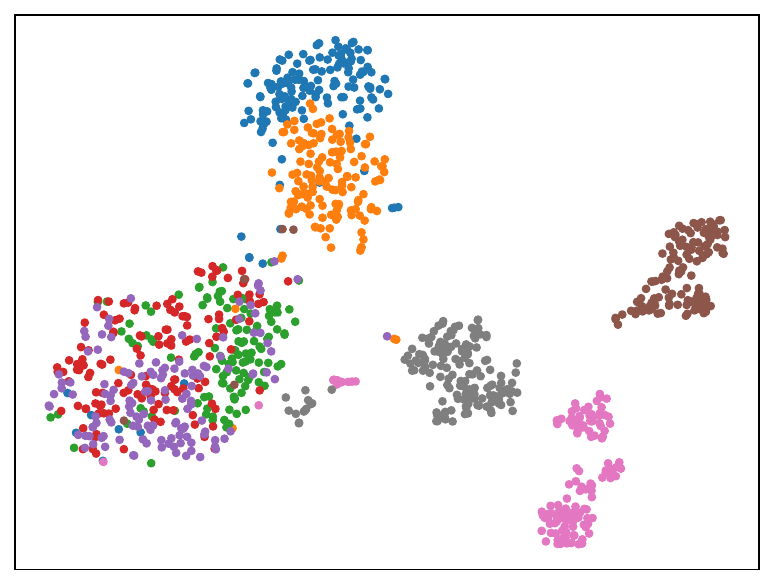}
            \caption{Low-light}
        \end{subfigure}
    \end{minipage}
    \caption{Visualization of appearance features of LMOT dataset using t-SNE. The same object is coded by the same color. It indicates that the appearance of objects under well-lit is distinguishable, but significantly reduced under low-light conditions}
    \label{fig:exp_analysis_tsne}
    \vspace{-1em}
\end{figure}

\noindent \textbf{Impact to association modules}. We analyze the impact of low-light conditions on the object association modules. Motion and appearance are important for object association. Both of them rely on detection boxes to locate the objects. To separate the impact of the detector, we use the ground-truth boxes as the detection boxes. The results are shown in \cref{tab:exp_analysis_mot}. It can be seen that using only appearance matching under well-lit conditions achieves the best result while using only appearance matching under low-lit conditions results in very poor performance. This indicates that the objects in LMOT have obvious visual distinguishability, but the distinguishability is significantly reduced under low-light conditions. In \cref{fig:exp_analysis_tsne}, we visualize the appearance feature of objects in the LMOT dataset under both well-lit and low-light conditions. We can observe that under well-lit conditions, LMOT is very distinguishable in the feature space. However, under low-light conditions, this discrimination of LMOT decreased significantly.

\subsection{Results on LMOT dataset}
We compare the proposed method with potential low-light multi-object tracking methods. 
We first train three baseline models using low-light videos, well-lit videos, and both of them, without any modifications to baseline trackers. They are denoted by Base-low, Base-well, and Base-all. The second type of method is tracking after low-light enhancement. These methods use low-light enhancement methods as pre-processing module. Both the enhanced videos and well-lit videos are used for training. We select four different low-light enhancement techniques for comparison. LLFlow \cite{wang2022low} and SDSD \cite{wang2021seeing} are low-light enhancement methods based on RGB images, while DNF \cite{jin2023dnf} and SMOID  \cite{jiang2019learning} are based on RAW low-light enhancement methods. LLFlow and DNF take images as input, SDSD and SMOID take videos as input.  The RAW object detection method is also trained together on well-lit and low-light videos. Its output bounding boxes are directly fed to the tracker. We test all the potential methods on two state-of-the-art trackers ByteTrack \cite{zhang2022bytetrack} and OC-SORT \cite{cao2023observation}. We also show mAP, AP50, and AP75 to highlight the detector performance. 

From  \cref{tab:exp_main_dual} we can see that, the proposed LTrack achieves the best HOTA and is highly competitive on all metrics. For example, the proposed method improves HOTA $1.3$ with almost the same parameters and computation as base methods. This strongly proves the effectiveness of the proposed method. Compared with tracking after low-light enhancement, these methods have a large amount of additional parameters and computation but still perform worse than the proposed method. For example, LLFlow delivers $8$ times FLOPS but still performs worse than our LTrack. In addition, We observed that there is no significant difference in results between RAW-based image enhancement methods and RGB-based image enhancement methods. This is not consistent with what our method has observed. The reason may be that low light enhancement focuses on image quality restoration and may mislead downstream tasks. As for RAOD \cite{xu2023toward}, it has almost the same number of parameters and computational load as our method, but its performance is much lower than the proposed LTrack. Despite addressing HDR scenes through a preprocessing module for RAW input, it does not perform well under low-light conditions.


\begin{table}
    \setlength{\abovecaptionskip}{2pt}
    \centering
    \resizebox{1\linewidth}{!}{
        \begin{tabular}{lccccc}
            \toprule
             Method & HOTA & AssA & IDF1 & MOTA & Det \\
            \midrule
            Base-low                       & 32.5 & 33.4 & 39.3 & 42.4 & 27.1 \\
            Base-well                       & 28.0 & 30.4 & 34.1 & 36.9 & 23.3 \\
            Base-all                        & \underline{34.6} & 38.1& \underline{40.0}  & \underline{44.8}  & \underline{30.2}\\
            \midrule
            LLFlow\cite{wang2022low}        & 33.6 & 38.0 & 38.4 & 43.1 & 29.7 \\
            SDSD\cite{wang2021seeing}       & 31.6 & 34.2 & 36.7 & 38.8 & 27.5 \\
            \midrule
            DNF\cite{jin2023dnf}            & 33.2 & \textbf{39.0} & 36.9 & 44.4 & 30.1 \\
            SMOID\cite{jiang2019learning}   & 31.2 & 34.1 & 36.6 & 41.2 & 26.9 \\
            \midrule
            RAOD\cite{xu2023toward}         & 32.2 & 33.5 & 39.4 & 40.9 & 26.5 \\
            \midrule            \rowcolor{gray!20} \textbf{LTrack (Ours)} & \textbf{35.1} & \underline{38.9} & \textbf{40.4} & \textbf{45.2} & \textbf{30.7} \\
            \bottomrule
        \end{tabular}
    }
    \caption{Experimental results on LMOT-real datase with OC-SORT\cite{cao2023observation}. The best result are shown in boldface}
    \label{tab:exp_main_real}
    \vspace{-1em}
\end{table}

\subsection{Results on Real World}
To validate the performance of our method in real-world low-light scenes at night, we evaluate all methods on LMOT-real dataset, using OC-SORT \cite{cao2023observation}. As shown in \cref{tab:exp_main_real}, the proposed LTrack performs much better than all comparison methods. Both the methods of tracking after low-light enhancement and the RAW detection method encounter generalizability problems and are not even better than Base-all. This strongly demonstrates the robustness of our LTrack in real-world dark scenes.

\begin{table}
    \setlength{\abovecaptionskip}{2pt}
    \setlength{\tabcolsep}{4pt}
    \centering
    \resizebox{1\linewidth}{!}{
    \begin{tabular}{cc|ccccc}
        \toprule
        \makebox[0.06\textwidth][c]{DSL} & \makebox[0.06\textwidth][c]{ALD}  & HOTA &  AssA & IDF1 & MOTA& DetA \\
        \midrule
        &                                 & 28.1 & 42.6 & 32.9 & 23.8  & 18.7 \\ 
        \checkmark &                      & 29.0 & 42.4 & 35.0 & 25.8 & 20.1  \\
         & \checkmark                     & 28.3 & 42.4 & 33.1 & 24.6 & 19.3 \\
        \rowcolor{gray!20} \checkmark & \checkmark & \textbf{29.4} & \textbf{43.2} & \textbf{35.2} & \textbf{26.1} & \textbf{20.3} \\
        \bottomrule
    \end{tabular}
    }
    \caption{Ablation on adaptive low-pass downsampling (ALD) and degradation suppression learning (DSL).}
    \label{tab:exp_ablation}
    \vspace{-0.5em}
\end{table}

\begin{table}
    \setlength{\abovecaptionskip}{2pt}
    \centering
    \resizebox{1\linewidth}{!}{
    \begin{tabular}{l|ccccc}
        \toprule
        Data Type & HOTA &  AssA & IDF1 & MOTA & DetA \\
        \midrule
         sRGB           & 29.2 & 42.7 & 34.9 & 26.4 & \textbf{20.5} \\
         RAW 8-bit      & 29.0 & 42.3 & 34.5 & 26.1 & 20.2 \\
         RAW 10-bit     & 29.3 & 42.9 & 34.8 & \textbf{26.5} & 20.4 \\
         \rowcolor{gray!20} RAW 12-bit  & \textbf{29.4} & \textbf{43.2} & \textbf{35.2} & 26.1 & 20.3 \\
        \bottomrule
    \end{tabular}
    }
    \caption{Results of different image format and dynamic range on LMOT test set.}
    \label{tab:exp_exp_nbit}
    \vspace{-0.5em}
\end{table}

\begin{table}
    \setlength{\abovecaptionskip}{2pt}
    \centering
    \setlength{\tabcolsep}{3.5pt}
    \resizebox{1\linewidth}{!}{
    \begin{tabular}{l|ccccc}
        \toprule
        Methods & HOTA & AssA & IDF1 & MOTA & DetA \\
        \midrule
        Gaussian-Possion      & 29.9 & 43.1 & 35.7 & 26.6 & 21.1  \\
        Physical-based \cite{wei2020physics}        & 30.4 & \textbf{44.1} & 36.0 & 26.6  & 21.2  \\
        \rowcolor{gray!20} \textbf{LMOT (Ours)} & \textbf{35.1} & 38.9 & \textbf{40.4} & \textbf{45.2} & \textbf{30.7} \\
        \bottomrule
    \end{tabular}
    }
    \caption{Results with low-light synthesis methods and LMOT dataset on LMOT-real set.}
    \label{tab:exp_lmot_sync}
    \vspace{-1em}
\end{table}

\subsection{Exploration and Discussion}
In this section, we conduct extensive analysis and discussion on LMOT datasets and the proposed method.

\vspace{0.5em}
\noindent \textbf{Ablation Study.}
We conduct ablation experiments to validate the effectiveness of our improvements, the results are shown in \cref{tab:exp_ablation}. It can be seen that all improvements contribute effectively to enhanced performance. Among them, the degradation suppression learning strategy demonstrates the most significant effect, resulting in an approximate $1$-point enhancement in HOTA. The best results are achieved when employing all strategies simultaneously. This demonstrates the effectiveness of all our contributions.

\vspace{0.5em}
\noindent \textbf{RAW \vs sRGB.}
We analyze the impact of input data formats, the results are shown in \cref{tab:exp_exp_nbit}. The 12-bit RAW format achieves significantly better results than sRGB. Because the RAW format saves much more potential information and is helpful for MOT in low-light scenes. We also observed that higher bitwidth is beneficial for performance, which has also been observed in other vision tasks \cite{hong2021crafting,xu2023toward}.

\vspace{0.5em}
\noindent \textbf{LMOT \vs Synthetic data.}
We compare our LMOT dataset with the synthetic low-light data to further demonstrate the value of LMOT. We apply the Gaussian-Possian based and Physical-based \cite{wei2020physics} low-light data synthesis method to synthesize low-light videos from well-lit videos. We train our LTrack on these types of data and evaluate their performance in real low-light scenes using LMOT-real. As shown in \cref{tab:exp_lmot_sync}, the tracker trained on our LMOT dataset has much better performance in real night scenes, which strongly demonstrates the value of our LMOT dataset.

\vspace{0.5em}
\noindent \textbf{Analysis on different category.}
We also analyze the performance for different categories. From \cref{tab:exp_exp_category}, we can see that cars and buses achieved the best performance because they have regular shapes and larger areas. Trucks exhibited the poorest performance, because they have the fewest instances, making it challenging for the model to learn accurate identification. The person achieves relatively average scores. Bicycles and motorcycles have close scores since they have similar appearance and motion patterns.

\begin{table}
    \setlength{\abovecaptionskip}{2pt}
    \centering
    \begin{tabular}{l|ccccc}
        \toprule
        Categories & HOTA & AssA & IDF1 & MOTA & DetA \\
        \midrule
        Person      & 24.2 & 30.8 & 29.9 & 25.1 & 19.2  \\
        Bicycle     & 16.9 & 40.3 & 15.6 & 8.9  & 7.1  \\
        Car         & 37.5 & 53.3 & 47.2 & 33.7 & 26.5  \\
        Motorcycle  & 19.8 & 35.0 & 22.6 & 15.2 & 11.4  \\
        Bus         & 44.3 & 59.9 & 51.8 & 36.7 & 32.9  \\
        Truck       & 6.9  & 15.6 & 6.4  & 3.2  & 3.1  \\
        \bottomrule
    \end{tabular}
    \caption{Results of the proposed method for different categories on LMOT test set.}
    \label{tab:exp_exp_category}
    \vspace{-1em}
\end{table}

\section{Conclusion}
\label{sec:conclusion}
In this work, we investigate the multi-object tracking in the dark scenes. We build a new low-light multi-object tracking (LMOT ) dataset, which provides well-aligned low-light video pairs and high-quality multi-object tracking annotations. We observed that low-light images are significantly degraded by the sensor noises, which also degrades the feature maps and significantly deteriorates the model performance. To learn the invariant semantic formation under noise disturbance and quality degradation, we present the adaptive low-pass downsample module and degradation suppression learning. These improvements greatly enhance the robustness of our method in real-world low-light scenes.

\noindent \textbf{Limitations.} We focus on multi-object tracking in the dark scenes. But we do not consider other degradation environments in the real world, such as rainy and foggy days. In our future work, we will consider exploring multi-object tracking in more real-world scenarios, promoting the development of MOT in real-world applications.

\section*{Acknowledgments}
This work was supported by the National Key R\&D Program  of China (2022YFC3300704), the National Natural Science Foundation of China (62331006,62171038, and 62088101), the R\&D Program of Beijing Municipal Education Commission (KZ202211417048), and the Fundamental Research Funds for the Central Universities.

{
    \small
    \bibliographystyle{ieeenat_fullname}
    \bibliography{main}
}


\end{document}